# Roadmap for using large language models (LLMs) to accelerate cross-disciplinary research with an example from computational biology

Ruian Ke[1,*] and Ruy M. Ribeiro[1,*]

[1] Theoretical Biology and Biophysics Group, Theoretical Division, Los Alamos National Laboratory, Los Alamos, NM 87545 USA

*Corresponding authors:

R.K.: rke@lanl.gov ; R.M.R.: ruy@lanl.gov

**Abstract**

Large language models (LLMs) are powerful artificial intelligence (AI) tools transforming how research is conducted. However, their use in research has been met with skepticism, due to concerns about hallucinations, biases and potential harms to research. These emphasize the importance of clearly understanding the strengths and weaknesses of LLMs to ensure their effective and responsible use. Here, we present a roadmap for integrating LLMs into cross-disciplinary research, where effective communication, knowledge transfer and collaboration across diverse fields are essential but often challenging. We examine the capabilities and limitations of LLMs and provide a detailed computational biology case study (on modeling HIV rebound dynamics) demonstrating how iterative interactions with an LLM (ChatGPT) can facilitate interdisciplinary collaboration and research. We argue that LLMs are best used as augmentative tools within a human-in-the-loop framework. Looking forward, we envisage that the responsible use of LLMs will enhance innovative cross-disciplinary research and substantially accelerate scientific discoveries.

# Introduction

Artificial intelligence (AI) has transformed many areas of scientific research [1]. For example, machine learning based models designed for processing and generating predictions from large scale datasets made many breakthrough discoveries that were not possible before [2]. More recently, large language models (LLMs) have emerged as a powerful AI technology, offering unprecedented capabilities in synthesizing knowledge and generating meaningful ideas in response to prompts using natural language [3]. This opens new avenues for integrating AI into broader day-to-day research activities beyond data modeling and prediction generation [4-11]. With the rapid improvements in LLMs' abilities, their impact on the future of scientific research will go far beyond their current usage.

Despite these exciting prospects, the use of LLMs in research have not been without criticism and concerns [12-15]. Some examples include the generation of inaccurate or misleading information, commonly referred to as "hallucinations" [13, 16], as well as the tendency of these models to oversimplify complex domain-specific nuances, and the potential for reinforcing biases embedded in their training data [12, 17]. These limitations raise questions about their reliability in scientific research. Based on our experience interacting with scientists across various disciplines, it is clear that opinions about the usefulness of LLMs are divided. Some believe that LLMs have the potential to excel in all aspects of research and thus fully adopt their outputs, while skeptics doubt the usefulness and accuracy of any responses they generate, thus completely reject their use in research. It is also pointed out that the coherent and fluent responses produced by LLMs can create a false sense of understanding, potentially diminishing researchers' grasp of fundamental scientific principles [12, 14]. Another criticism is that overreliance on LLMs may make science less innovative and more vulnerable to errors and lead to erosion of trust of scientific research [18] and homogenization of research approaches and perspectives [14].

Against this backdrop, we view LLMs as both a transformative and a potentially risky tool (as any new technology) available for researchers and seek to provide a balanced analysis of where (current) LLMs excel and where they fall short and need careful oversight within the research process. By examining LLM responses during each stage of research, from literature review to methodology design, data analysis, and drafting of manuscripts, we aim to equip researchers with a better understanding of how to think about LLMs' responses critically, use LLMs responsibly, and thus better integrate LLMs into their research workflows. Previous works have focused on general principles and benefits of using LLMs to conduct research, get grants, and assist education etc. [5, 8, 9, 19-21]. Here, we particularly focus on using LLMs for cross-disciplinary research, because many breakthroughs of scientific research and major innovations in technology arise from cross-disciplinary collaboration [22-24] and in our opinion, this is one of the areas where LLMs are most useful (**Figure 1**). We illustrate how LLMs can help facilitate cross-disciplinary collaboration, aiming to offer a useful introduction for researchers beginning to integrate these tools into their work. We also provide a practical end-to-end example from computational biology (i.e. modeling HIV rebound dynamics) to illustrate how one may interact with LLMs (in particular ChatGPT) iteratively in an actual project. At the same time, by highlighting current limitations, we emphasize the areas where responses generated by LLMs require careful examination and validation from experts (**Figure 1**). Finally, we acknowledge that although the roadmap we provide here may be applicable to current models, the capabilities of LLMs are rapidly improving. Thus, aspects of this roadmap may need revision as technology unfolds.

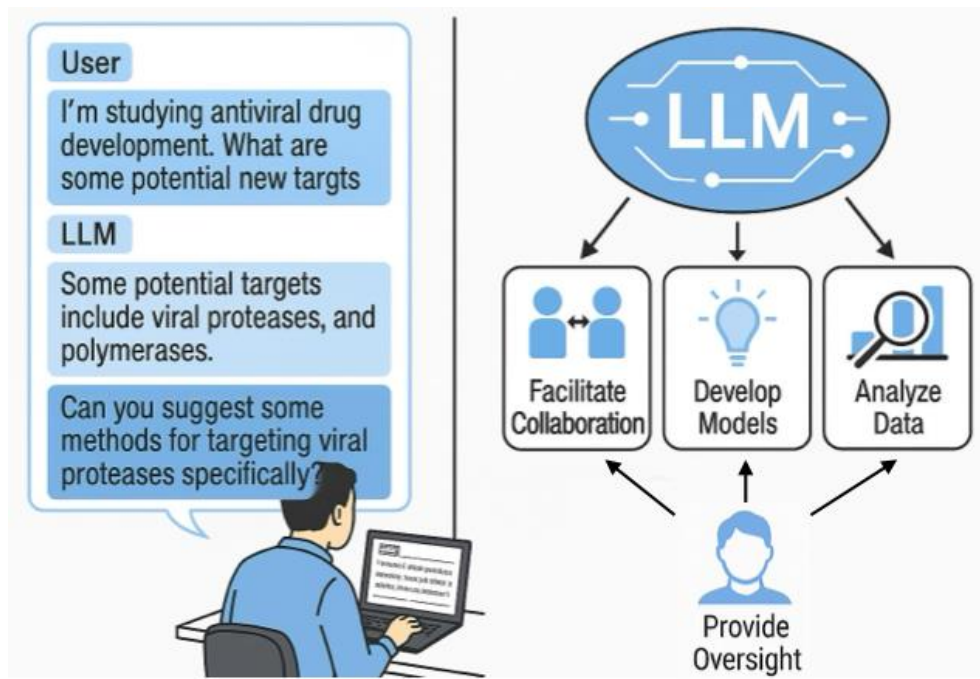

**Figure 1. Integration of LLMs into cross-disciplinary research workflows**. Effective and responsible use of LLMs with domain expert oversight facilitates communication among researchers from diverse scientific disciplines, guides the selection of suitable computational and statistical methods to analyze heterogeneous datasets, and ultimately accelerates scientific discovery. The figure is adapted from an image generated by ChatGPT 4o using the 'Create image' feature.

**Using LLMs to Navigate and Connect Diverse Scientific Literature**
A literature review, for cross-disciplinary research in particular, involves an extensive examination of scholarly articles and books across diverse fields. This process is not only labor-intensive but also sometimes it is hard or impossible for researchers from one field to read through papers of other fields to fully grasp the technical details. Thus, a common challenge in cross-disciplinary collaboration is to understand the fundamentals of multiple fields, so that researchers can communicate effectively by speaking and understanding each other's 'language'. Much effort and energy in cross-disciplinary collaboration is therefore devoted to clarifying technical details. This can be a major factor hindering cross-disciplinary collaborations.

Several capabilities make LLMs particularly useful for addressing this problem: 1) LLMs are typically trained on an enormous amount of text, including diverse research literature, 2) current

LLMs are equipped with the capability to search the internet and retrieve and analyze new data sources, and 3) they are capable of translating complex domain-specific jargon to plain and easily accessible language and present information in a well-structured way. The quality of LLM response is heavily dependent on the prompts used to query the LLM, i.e. so-called 'prompt engineering'. To obtain a good response, one important aspect is to provide sufficient context such as what one's expertise and specific needs are and what one is trying to achieve. See Category A in Table 1 for some examples of general prompts for literature research. With appropriate prompt instruction, they can analyze and summarize existing literature in a specific field within minutes. The responses usually provide both concise summary and comprehensive coverage of the research themes. One can also further query LLMs about specific technical terms or methods for more detailed descriptions. We found that in general, the background information in the LLM responses provides researchers who are new to a field with a very good overview of the major research themes and questions and can effectively facilitate smoother interdisciplinary communication.

If one would like to synthesize research literature across many disciplines, one could query LLMs to provide a summary across the different fields. A potentially useful feature is that when specifically asked to brainstorm a list of ideas that connect different fields of interests (see Category A in Table 1), LLMs may provide ideas on the technical or conceptual connections between the different fields of interests. In our experience, LLMs sometimes are capable of producing interesting insights—many of which appear to be genuinely novel and not yet explored in the literature. However, we caution that the list may contain many infeasible or illogical ideas, although their explanation often sounds reasonable at the surface level. Expert evaluation remains essential to assess the feasibility and relevance of these ideas and to prioritize the most promising directions.

A common concern, especially for literature reviews, is the issue of "hallucination", e.g. generating non-existent or inaccurate references, or statements not actually found in the cited sources. This issue can be more problematic in interdisciplinary research, when one does not know the literature well. To address this issue, ChatGPT and other LLMs have introduced features like the "deep research" functionality, which appears more effective at summarizing and synthesizing research findings and accurately citing original sources. This feature largely mitigates previous model limitations, such as 'outdated training data' and 'hallucination'. Therefore, we recommend using this functionality whenever a review or search of the literature is needed. Moreover, one can request the LLM to retrieve the actual study (e.g, from PubMed).

Overall, effective use of LLMs can reduce the time and effort required to build an understanding of the existing literature, and their ability to integrate and translate information across disciplines makes them particularly valuable in cross-disciplinary research contexts. On the other hand, we would like to emphasize that their use comes with important caveats and risks and requires careful examination. First, although they generally offer a comprehensive overview of a research topic, these models, including the 'Deep research' function, may overlook nuanced arguments or fail to fully capture the complexity of individual studies. Human judgment is critical in verifying that the synthesized information accurately represents the content and context of the original studies. Second, LLMs can inadvertently perpetuate biases present in the literature or produce overly generalized conclusions that require critical evaluation. We should remain aware of this risk, and carefully consider the representativeness of the presented information. Third, even with 'Deep Research'-like capability, 'hallucination' may occur sometimes. It remains important to cross-check LLM-generated citations and summaries against original primary sources and domain experts. Using information from LLM responses without human (expert) in the loop supervision is definitely not advised. Ultimately, checking references and their content remains the responsibility of the researcher, as has always been the case.

**Example:**
We will use a specific example project throughout this manuscript to demonstrate how one in practice interacts with LLMs in a single thread to effectively use them as assistants to accelerate research. We included the prompts we used, the ChatGPT responses and the evaluation of the ChatGPT responses to illustrate how one can query LLMs and critically assess and use the responses to effectively assist a multi-disciplinary project.

In this project, we built and fit mathematical models to clinical data of HIV rebound dynamics after analytical treatment interruption (i.e. intentional treatment cessation in participants who previously were effectively treated with antiretroviral therapies). The aim is to understand the underlying biological driver(s) of the different patterns of the viral load rebound dynamics observed in different individuals. This work was originally performed without LLM assistance and published previously[25]. Here, we assumed that this work was not done and instructed ChatGPT to avoid using this paper in its searches and responses.

First, we would like to review the literature on the biological and clinical findings about HIV rebound. We entered the prompt below and used the "Deep Research" functionality in ChatGPT:

```
'I am working on understanding HIV viral load rebound after ATI. Can you
   me[a] a summary of research on the current hypothesis of why and how the
          rebound occurs, and the different outcomes of ATI?'
```

The LLM asked a few clarifying questions (e.g., should it concentrate on human studies or include non-human primate studies? should it focus on recent studies from the last 5 years? etc). This is a hallmark of the function of the "Deep Research" feature. Answering these questions allowed it to focus on a more relevant answer. After about 7-10 minutes, we got a comprehensive summary of the literature which was divided into three major themes: mechanistic insights, clinical outcomes and human vs. non-human primate studies (see Supplementary Information). We feel that the response provides a very good overview of the field, highlighting the key questions and themes.

---

[a] Here we intended to type in 'Can you give me …' instead of 'Can you me…'. Nonetheless, the typo did not affect the response.

It also provided, of its own volition, a table summarizing "ATI Outcomes and Predictive Factors (2019-2024)", and listing studies and their findings according to the clinical outcomes. These are very useful and would take a researcher hours/days to collect. We did note that for certain passages, it cited the same reference over and over, and sometimes even repeated the same reference twice in a row.

We are more interested in the literature for mathematical modeling of HIV rebound, and thus followed up with the prompt using the 'Deep Research' function:
```
'I'd like to work on mathematical modeling of viral rebound. Can you tell
   me about related works and relate that to the studies you mentioned
                              above? '
```
Again, after a few clarifying questions, ChatGPT responded with a comprehensive summary organized into several major themes. It starts with mathematical modeling approaches to HIV rebound and their applications to data, and then focuses on topics, such as reservoirs dynamics, immune control during rebound, predicting time-to-rebound and modeling intervention strategies. It covers a wide range of topics and most of the information was correct.

Overall, the responses generated by 'Deep Research' was very useful, with correct information and good summaries that relied on not only key papers, but also on conference abstracts and news pieces from HIV information sites (e.g., linked with clinical trials' press releases). The summary was an excellent introduction to the theme (serving a useful material for researchers who are new to a field) and very relevant to our objectives, even if one needs to be careful to corroborate the information provided. However, we feel that it lacks nuanced discussion of the major barriers, unknowns or limiting factors in each subfield. For the purpose of identifying a novel research question, the high-level summary may be insufficient.

Next, we were curious if what we were planning to do potentially would allow novel insights (note that we specifically asked ChatGPT to avoid using any information from our published work on this topic [25]), and posed this question directly with the prompt:
```
'I am trying to construct mathematical models of HIV rebound and fit the
    model to human datasets (longitudinal viral load measurements). The
      dataset includes VL data from a group of PTC and a group of non-
   controllers. I would like to estimate the parameter values and figure out
      the difference in parameter value distributions between PTC vs. non-
          controllers. Will we learn something new from this proposed approach?'
```
The answer generated summarized neatly our own thoughts on the subject, and the main reasons why modeling this phenomenon is relevant. In particular, the section on "Novelty and Advantages of Your Approach" lists reasons that align very well with our thinking when we conducted this study. The section on "Potential Challenges to Consider" (model complexity vs. identifiability; data quality; and visualization) was spot on. The 'Recommended Steps to Maximize Insights' provides very reasonable, albeit general, model development strategies to solve a complex problem.

To sum up our evaluation, the LLM responses provide a helpful overview of a research field by breaking the topic down into several subfields. Often the papers identified by LLMs are seminal or important research studies within those subfields, while other times the suggestions may be selected primarily due to their relevance to the user's initial prompt. Nonetheless, the responses should serve as a good starting point for an entry-level PhD student or a researcher, who is new to a field to get a good general understanding of a topic. It is also a helpful refresher for more experienced researchers. Of course, a nuanced understanding requires further iteration with LLMs, reading and reviewing specific papers and interaction with human experts.

**From Preprocessing to Plotting: using LLMs for Cross-disciplinary Data Analysis**

Proper cleaning and visualization of datasets are crucial to get an intuitive understanding of the patterns and trends in the data such that appropriate research strategies and approaches are designed to test hypotheses using these datasets. Irrespective of the size of the datasets, cleaning data to a machine-readable format is often a laborious and time-consuming task. Researchers often deal with missing values, outliers, or noisy data, using various techniques. With proper description of the dataset and the goal of data preprocessing in prompts (see Category B in Table 1 for examples), LLMs can effectively generate code that automates data cleaning tasks. Although working iteratively with LLMs may be needed to produce codes that fully process complex datasets without error, this capability of LLMs significantly reduces the time spent on preliminary data preparation.

A hallmark of modern cross-disciplinary research is the collection and availability of various types of datasets. Sometimes statistical or computational experts are not fully aware of all the subtleties of the data and the context of the specific research field. The ability of LLMs to suggest statistical and computational methods in the context of a research domain and provide corresponding codes (in a variety of programming languages) is one of the greatest strengths of LLMs for research. LLMs can relate the research literature from the field of interests to the computational or statistical task in hand and thus compare the strengths and limitations of various methods based on the nature of the data and the research questions. Sometimes it can even suggest hybrid approaches that might combine multiple methods.

Data visualization also plays an essential role in helping communicate findings effectively, especially in a multi-disciplinary team. However, graphing professionally looking plots in programming languages often requires comprehensive knowledge of plotting packages/software and extensive trial and error. In this context, LLMs are great tools that can be used to generate descriptive statistics, suggesting appropriate visualizations—such as histograms, scatter plots, box plots, and more advanced graphs — and generate the code required to produce them (see Table 1). Again, the practice of human (expert) in the loop in this stage is essential, i.e. always ask LLMs to provide the source code used for data cleaning and visualization to allow for a human expert to ensure correctness.

After visualization, LLMs can be used to read graphs, interpret and reason with data and statistical results. These allow them to provide context by relating the results to established theories or by highlighting unexpected patterns that merit further investigation (see Category B in Table 1 for example prompts). Of course, this capability depends on the type of data and graphs. We found, given sufficient context of the data, LLMs can interpret a figure well when the raw data in numerical format is available. When they are asked to read a figure directly from a graphic file, they may or may not interpret the graph correctly. We recommend carefully examining the responses from LLMs, for example, one could also ask the model to state what information it gets from the graph to ensure the model reads the graphs correctly.

**Example**

In our example (Supplementary Material), we collected a set of anonymous clinical data of longitudinal HIV load measurements after analytical treatment interruption. We would like to first perform visualization and statistical analyses of key features of this dataset, and then perform statistical analyses of the data to identify potential differences between the two groups of individuals, where their clinical outcomes differ, i.e. the post-treatment controller vs. the non-controller group.

First, we prompted ChatGPT to visualize the data:
> 'The attached file contains longitudinal viral load data from each individual (ID is in the first column). The second column contains the duration of infection, and the third column has the outcome after ATI. The remaining columns are viral load by week post ATI (as column names). Note when viral load is at 50, it means undetectable viral load (under the limit of detection, i.e. 50 copies/mL). Can you read the data in and visualize the viral load dynamics and group them according to the outcome?'
> 'Can you plot the viral load on a log10 scale? and show the individualized viral load? it'd be good to use colors for the two groups with a larger contrast. '

The visualization is excellent for data inspection, although some further refinements would be needed to make the figure publication ready.

We then asked it to calculate summary statistics and to suggest appropriate statistical tests to compare the longitudinal viral load data from the two groups of individuals:
> 'Can you calculate summary statistics such as peak viral loads, time to rebound (defined as above the limit of detection, i.e. 50 copies/ml), and set-point viral load (defined as long term viral load) and compare these between the PTC and NC groups, and visualize the results?'

```
'Can you suggest appropriate statistical tests to test the difference
     between the two groups, perform them and return the results?'
```
The response from LLMs is accurate. The statistical test used was the one we would have chosen (the Mann-Whitney) for this purpose. We checked the p-values, and they were also correct. Note that the LLM used the absolute values for viral load, rather than $\log_{10}$, but this is not relevant for the non-parametric test, which only uses ranks. When we asked to visualize the differences assessed with the statistical tests, it correctly chose a box-plot representation. However, it made the mistake of putting all the box plots in the same figure frame, which mixes time variable with viral load variables. In addition, the best-practice for boxplot is to overlay the individual data points on top of the box-plot, especially for a small number of samples. Further queries are needed to improve these plots.

A very important step in data visualization (or any task involving computation) is to ask LLMs to provide all the code used, because it enhances reproducibility and allows us to customize any details needed. We thus asked ChatGPT to provide all codes:
```
      'Can you show me the code that performed all the analyses and
                            visualizations?'
```

**Assisting Methodology and Model Development with LLMs**

Once the research question is established, researchers in a team typically collaborate to formulate a general methodological framework for conducting experiments, analyzing data and developing models. At this stage, LLMs will be very useful to clarify the details in the choice of methods under each specific step. For instance, when provided with a clear description of the dataset and the research objectives in a prompt (see Category C in Table 1), LLMs can suggest a list of potentially feasible experimental or computational approaches, along with pros and cons of these approaches. In our experience, the responses usually provide reasonable summaries and suggestions, although it remains unreliable to determine the best approach simply based on the summaries. These summaries should serve as complementary information, alongside standard practices in the field and specific nuances of the research context, to help researchers determine the most suitable approach (also see the example below).

Large-scale and heterogeneous datasets become increasingly common in experimental, clinical and field studies, and ML or computer simulation approaches become increasingly popular for identifying hidden patterns in large-scale datasets. However, due to the large variety of models and the complexity in implementing them, there is a high barrier even for quantitative researchers to implement different ML methods. For example, in one of our projects where we tested different types of neural network architectures (e.g. long short-term memory, Transformers, etc.)[26], one of us with a computer science background spent months correctly implementing and validating these models against the datasets. With LLMs, the barrier to entry and the time cost would be largely reduced by querying LLMs to compare different ML approaches, generate codes for models with different complexity (see example prompts in Table 2). More broadly, we find that generating codes for data analysis, visualization and computational model development in a variety of programming languages, such as Python, R, etc. is one of the most powerful capabilities of LLMs (Table 2). This substantially decreases the time and effort that researchers devote to coding.

Another challenge is related to the high complexity and computational costs to test a variety of candidate methods. In this case, using an incremental procedure involving iterative interaction with LLMs can drastically reduce the time and cost. We recommend starting from querying LLMs to generate simplified 'toy models', which can then be tested on smaller but representative subsets of the dataset. Once a method is confirmed to be feasible through initial tests, researchers can iteratively refine and expand upon these toy models through further interactions with LLMs to develop fully functional models capable of analyzing the complete dataset. This incremental approach helps decreases the complexity and computational cost of testing large models and minimize the risk of errors in the LLM generated code.

There are, of course, limitations. First, although LLMs can make suggestions and summarize advantages/disadvantages of different methods, eventual selection of an approach still needs substantial expertise and careful consideration to ensure that the selected approach is both feasible and scientifically sound. Perhaps this is the critical step of the scientific process, which takes a researcher many years to hone, and cannot yet be replaced by an LLM. A brute force approach, for example trying to use all the methods proposed, is clearly not advisable. Second, although the code provided by LLMs may run without issues, coding expertise and understanding what the code represents in relation to the actual underlying process are critical to validate (and potentially debug) the provided code (in a local programming environment), instead of assuming the correctness of the code without checking. In this regard, LLMs ability to carefully annotate the code is an important asset. Moreover, one can ask the LLM (or another LLM) to check the code. AI agents are already developed to run and correct the code iteratively and autonomously[27]. However, errors and discrepancies between intended and coded mechanisms/algorithms often still exist because of error in the generated code or misunderstanding of or ambiguity in the prompts. This is why we recommend an incremental approach where one can break down complex algorithms into clear instructions of each step. Nonetheless, an initial code draft using appropriate packages with minimal errors saves much time compared to coding from scratch.

**Example:**
In our example, after the literature review and data visualization, we would like to build a mechanistic model to understand the processes underlying different outcomes, i.e. post-treatment control (PTC) or non-control. We wanted to explore possible mathematical modeling approaches that would allow us to use the data to test different hypotheses contributing to the different outcomes (as mentioned above, we already had a clear idea of what we wanted to do, nevertheless this more phased approach to using the LLMs can be insightful). We used the following prompt:

```
'I have the dataset of longitudinal HIV load measured from more
    than 20 individuals who underwent analytical treatment
```

> interruption. Given what we discussed, can you suggest modeling
> approaches to analyze the data to get new insights on PTC? I am
> particularly interested in the mechanisms of PTC, and distinguishing the
> hypotheses of the determinants of PTC: size of reservoir vs. immune
> control?'

The LLM response (Supplementary Material) included an overview as well as details of each step, i.e. sections on model structure, parameter estimation, hypothesis testing, possible insights and model extensions. Overall, the response provided a good general framework and aligned well with our planned approach. However, we feel that some of the descriptions were too general and did not have enough details for each step of research. For example, the "Data fitting and parameter estimation" section was quite minimal. No doubt follow-up questions would have improved these.

It is interesting that the LLM suggested an actual ODE-based model. This model was reasonable, but not fully correct, and an attentive eye is needed to correct it. Indeed, some terms in the deterministic model the LLM provided are not consistent with biological knowledge. We thus entered prompts to correct those inconsistencies:

> 'For the ODE model you provided, you missed the ODE for the target cell
> population. In addition, for the ODE for the latent reservoir (L),
> latently infected cells can proliferate and die (at slow rates), they are
> formed by a small fraction of viral infections of target cells. Can you
> revise the model?'

This was very detailed information that a domain expert would provide. These inputs allowed the LLM to immediately correct the equations and present the full mathematical system in a version nearly identical to the one we used in our recent paper[25].

With the model suggested by ChatGPT, we then would like to search the literature to figure out what is already known about the parameter values for the model. This will help us to determine the fixed parameter values and the parameters to fit to the datasets. Thus, we entered the prompt with the 'Deep Research' function:

> 'Can you find out the parameter values for this model from the research
> literature?'

The response provided a very good summary of the parameter values in a suitable table format. In the table, the response included not only references to relevant studies, but also a rationale of why each value was suggested. This response saves the enormous amount of time needed to go through literature to find relevant parameter values. Nonetheless, it is always key to evaluate the output carefully. In this case, we note that some of the references cited were not the original studies that estimated the parameter values. Of course, it is easy to read the cited paper and trace the original study.

Next, we would like the model to suggest candidate parameters to fit to the data:

> 'I would like to fit a subset of parameters in this model to the data I
> attached earlier. Can you suggest parameters to estimate?'

The response seems logical and consistent with the objectives of our project. It is also relevant that it includes reasons why the suggested parameters make sense, and also what parameters to keep fixed and why. However, there are nuances that the LLM did not consider, for example, we know that this type of data contains information more suitable to estimate some parameters than others, or that some parameters are correlated. This is another situation emphasizing the importance of human expert knowledge and judgement.

We thus were more direct and told the model which parameters to fit based on our past experience:
```
'I would like to estimate the following parameters: beta, p, delta0, and
all the parameters in the immune effector cell equation. Can you provide
a python code and a R code to read the data I attached and then estimate
   the parameters I mentioned and fix the other parameter to values you
    suggested in the table? please also include visualization showing the
                            fitting results. '
```
We were probably overly ambitious here, asking to perform several tasks in a single prompt, still the LLM generated the Python code and R code to generate the figures from the results. As usual, the code provided is well annotated, which makes it very easy to follow the logic. This is particularly useful when one uses a programming language for a specific purpose or translating an existing code to another language to increase computational efficiency or for easier use by another team member.

We were more interested in a mixed-effects approach, i.e. fitting a single model to data across a population simultaneously by estimating the distributions of parameter values[28]. In our field, we typically use the software Monolix (Lixoft, Antony, France)[28]. Since this is proprietary software and likely has a much smaller user base than Stan, this would be a more stringent example. We did this in multiple steps, asking:
```
 'I would like to fit the model (with the same parameter setting above) to
 data across all individuals using Monolix. Can you provide a Monolix file to
                               do so?'
```
Unfortunately, the code the model provided had many errors (possibly because documentation and explanation for Monolix and its language mlxtran are not as abundant as for e.g., python in ChatGPT's training dataset). We thus tested if ChatGPT is able to improve the code, if we provide an example file (that contains mlxtran and ODEs from a different project). This is a typical example of one-shot learning, at which LLMs are very good in general[3].

We uploaded the files in turn and asked:
```
 'Your mlxtran code has some problems. I attached an example code. Can you
  use this file as an example and combine what you know about mlxtran code
             to revise the mlxtran code for the HIV model?'
```
The re-worked code was much better, and we could make it work with just a few more tweaks. On the positive side, we were able to generate this code quicker than if we had coded it from scratch, and it will serve as template for other projects; on the other hand, good knowledge of the different aspects was essential for debugging.

In our research, it is important to test different hypotheses and corresponding models for their ability to explain a dataset and to make more robust conclusions about the underlying biological mechanisms that may control the viral dynamics. Thus, we asked ChatGPT to provide several possible extensions based on the research literature by entering prompts:

> 'For my research, I would like to explore different models based on different hypotheses regarding the role of the immune system, and then test which model fits the data better to find out the most important immune factor. Can you make a few suggestions how I could revise the model according to known mechanisms of how the innate and/or adaptive immunity impact on viral dynamics?'

The LLM provided a very worthwhile response, suggesting 5 different models including innate mechanisms (prototype interferon-based), antibody effects, and variants on cytolytic activity (based on CD8+ T cells or NK cells). Both the mechanisms and the functional forms proposed were reasonable, and easily testable. It also proposed a practical implementation strategy and how to compare the results of the different models, which follow standard practices. Overall, this was a very useful response that could be the basis for further exploration.

## LLMs for Drafting, Polishing, Bridging Disciplinary Languages and Reducing Human Language Barriers

One of the immediate benefits of using LLMs is in providing a first draft once appropriate context (e.g. research background, motivation and results) are provided. This substantially helps with avoiding writer's block faced by a lot of researchers when they start to write a paper from scratch. After interacting with LLMs in the research stages above, one can use the prompt shown in Category D in Table 1 to start a draft. To obtain more tailored responses, one should provide as much background and instruction for each section of the draft as possible. This could be an iterative process where a user provides feedback to each section of the draft and more specific suggestions for revisions.

Despite the benefit, we found that the text generated by LLMs is often very general, lacking specificity and ingenuity, and repetitive. This is possibly a consequence that LLMs are fundamentally a pattern matching tool, generating text based on statistical relationships found in the training data. This means that while they can produce coherent and stylistically appropriate text, this text may be not specific enough to describe the research (which by-definition is novel). One would need to revise the initial draft extensively to reflect the thinking process and the research procedures and findings appropriately. That is to say, the writer has to dictate the main content of a paper [29-31]. In this way, the paper truly reflects the innovative ideas, approaches and results of a team to advance our scientific knowledge. We do not recommend using the text generated by LLM directly without revision. Indeed, concerns in the ethics of using LLM generated content in academic papers are raised[6, 12, 13].

Once the content is written, LLMs can be a great tool to assist in improving the clarity and coherence of the text. This is a notable benefit in cross-disciplinary research contexts, where papers often integrate text written by team members from different disciplines and language and stylistic conventions can vary between fields. LLMs can help ensure that the manuscript adheres to the appropriate disciplinary norms and at the same time, can be understood by a broad spectrum of audience. Additionally, LLMs are particularly useful for non-native speakers to more accurately express their ideas, by suggesting rephrasing of awkward sentences, ensuring consistency in terminology, and helping to align the language with the formal tone typically required in academic publications (Table 1).

Overall, LLMs can be useful tools at the stages of drafting and final language polishing. However, the main content should be (mostly) written by human experts to faithfully reflect the research. Over-reliance on LLMs raises concerns related to originality and potential plagiarism[12, 13]. Finally, the use of LLMs in drafting and review processes must be transparent. Scholars should acknowledge the use of such tools in their methodology.

**Example**
Continuing our example (Supplementary Materials), we first asked ChatGPT (version 4.5) to write an introduction section, based on the previous discussion.

```
'Based on what we discussed and what I am trying to do, can you provide an
                    introduction for a research paper? '
```

We note that providing context for LLMs, such as a prompt like 'Based on what we discussed and what I am trying to do', is important for appropriate framing of the response. We found the response provided a very good introduction, which aligns well with professional standards both in terms of the style of the language and content, as well as length. The LLM divided the introduction into four paragraphs: i) a general introduction to ATI, viral rebound and post-treatment control and highlighting the importance of understanding the underlying drivers of the observed clinical outcomes - a goal of our study. ii) a presentation of the hypotheses about the drivers of control (reservoir-driven vs. immune-driven); iii) a discussion as to why mathematical modeling is suitable for analyzing the hypotheses; and iv) a final paragraph summarizing what is done in the present study. This is very similar to an introduction that we could have written ourselves.

We then asked ChatGPT to write the method section:

```
'Can you write a method section including the statistical analysis you did
  earlier and description of the mathematical models. For the mathematical
 models, let's assume we constructed and tested the baseline model and 3 of
  the candidate model extensions you suggested, and compared the models using
                    model comparison techniques.'
```

The response includes detailed description of the dataset, the statistical analyses and the mathematical models in a professional tone. The LLM performed a reasonable job, however, it presented them in summary form (e.g., using bullet points) and in somewhat general terms. Thus, although it is a good first pass, more work (e.g. correcting experimental or modeling details) needs to be done, either working with the LLM to refine this or manually starting from this draft.

We moved on to ask ChatGPT to write the Results section:

```
'I am writing the Results section for the paper. The results section should be
 structured like this: 1. Showing the viral load trajectories of the data and
   Analyze the data using simple summary statistics (such as those you have
 done). 2. Develop a baseline mathematical model to understand the drivers of
    post-treatment control, and show the results of model fitting. 3. Explore
other hypotheses about post-treatment control by extending the baseline model
 (as what you suggested). What we found in our analyses of the model is that a
    key parameter separating the dynamics of PTC vs. NC is the half-saturation
     term, h, in the model. Please write the Results to reflect that and please
           include all the graphs you generated in our conversation.'
```

Again, since most of the results are collected outside of ChatGPT, we start the conversation by explaining what results we would like to include and what the results are. ChatGPT described the results very well, especially for the statistical analyses where these results were generated earlier in the conversation. In particular, it was able to refer to the figures it had generated in the proper context. For the results that were not generated with ChatGPT, it managed to lay out a reasonable argument and discussion; however, the details of the actual results, which we did not provide, are made up by the model. Another issue is that the LLM repeated parts of the Methods, which usually would not be included in the Results section. This again indicates that careful revision of this section is important, especially to verify that the quoted results are indeed real.

Finally, we asked ChatGPT to write a Discussion section:

```
'Based on the results and all the conversations we had so far, can you draft a
   Discussion section for the paper? The Discussion should first emphasize the
     most important findings of this study in 3-5 sentences. Then, it should
  discuss each item of the results in the context of the literature. It should
       end with discussions of study limitations and possible future works.'
```

The response starts with a summary of the important findings of the study (as instructed). It then went into detailed discussion of the results (here we are assuming the results are real) from each analysis and how they fit with the broader research literature. We were impressed by how well it understands the results and manages to link them to existing studies. In addition, the discussion of the limitations of the study and future work are well constructed and fit into the larger research context. However, we felt it lacked some personal perspective that only an engaged researcher or team may be able to provide.

## Conclusion and Outlook

The integration of Large Language Models (LLMs) into cross-disciplinary research is reshaping how researchers approach and perform scientific tasks. LLMs can now effectively support researchers by handling various routine technical tasks and bridging knowledge gaps and facilitating interdisciplinary collaboration. By alleviating the burden of routine tasks, LLMs lower the barriers in cross-disciplinary research and enable researchers to concentrate more on aspects of research that demand creative and critical thinking, with the potential to accelerate discovery and innovation.

Looking forward, we anticipate increased integration of LLMs across various research tasks, for example, as demonstrated in Ref. [32]. As these models continue to evolve, the risks we touch upon above, such as 'hallucination' and generating errors in computer code, will be further mitigated. With more recent advances, e.g. in agentic system design, we expect more and more routine tasks (and sometimes creative tasks, such as idea generation and research planning) will be automated by LLM agentic systems (i.e. teams of LLMs) [1, 32-35]. We also expect LLMs will play a more and more important role in cross-disciplinary research. For example, LLMs may soon propose and implement effective methodologies that integrate and interpret diverse types of data—including text, image, and numeric datasets—enabling unified analyses across disciplines. With improved deeper contextual understanding, future LLMs could leverage their extensive training to identify novel interdisciplinary hypotheses and recognize critical knowledge gaps by synthesizing insights from diverse fields. Another fruitful area of LLM integration is in the training of the next generation of cross-disciplinary scientists, which is as crucial as advancing research itself. LLMs can offer personalized, adaptive educational experiences that efficiently teach key concepts and methodologies across different disciplines, preparing students to navigate interdisciplinary research effectively. This represents a uniquely valuable educational approach that traditional teaching methods do not offer.

While LLMs present powerful tools for enhancing productivity, we emphasize their role should remain complementary to human expertise. Rather than replacing human judgment, LLMs best serve as an assistant to human researchers to augment the critical and creative thinking that lies at the center of scientific inquiry. As we try to emphasize throughout the article, human (expert) supervision (i.e. the human-in-the-loop workflow) is critically needed to check the accuracy of LLM responses at each stage of LLM integration. Designing policies ensuring human oversight and transparency of the automated process is an emerging area that requires careful thinking and discussion [13].

**Acknowledgements**:

We thank Emma Goldberg and many scientists at Los Alamos National Laboratory for helpful discussions. The work was supported by NIH grants R01-AI152703 (RK) and UM1AI164561 (RMR).

**Competing interests:**

The authors declare no competing interests.

**Table 1. Tasks where LLMs can assist during each research step and suggested prompts.**

| Category | Tasks | Example Prompts* |
|---|---|---|
| A. Literature Review and Idea Generation | A1. General summary of a research field | I am conducting research on []. Can you summarize the research literature on []? Please list the links or citations of your referenced sources. |
| | A2. Summary of specific papers | I am conducting research on []. I have attached several research articles. Can you give me a summary of the major findings from these articles? |
| | A3. Organize literature into categories | Can you organize the literature on [] and the attached papers into different categories? |
| | A4. Draft structured literature review | Can you provide a draft literature review in approximately [how many] words based on the categorization above? Please include all citations and write the review geared towards a [] audience? |
| | A5. Understand an unfamiliar experimental approach | I am a mathematical modeler. I would like to know more about [an experimental approach]. Can you summarize the steps of this approach and explain it in plain language that a non-expert can understand? Please compare this approach to other similar approaches used to measure []. |
| | A6. Generate cross-disciplinary research ideas | I am doing research on solving a problem on []. I would like to draw ideas from two fields, i.e. [] and []. Propose a list of ideas that combine theories and [computational or experimental] techniques or draw analogies from these fields to solve the problem? Can you tell me the pros and cons of these different ideas? |
| B. Data Analysis, Visualization and Model Development | B1. Clean and load data with missing values | My data is attached as a csv file. The data is between column C and H and between line 3 to line 200. There is missing data in the dataset. Can you write an R script to read in the dataset? |
| | B2. Run statistical test and visualize results | The second method, i.e. [], you suggested is great. Can you provide R code to read the data from the csv file and perform the test you suggested? Please also visualize the results by comparing the two different groups using different colors. |
| | B3. Interpret and describe results from a graph | Can you analyze and interpret the results from the attached graph? And draft a description of the results. |
| C. Method Selection & Model Development | C1. Compare statistical methods and select appropriate analysis | I have a dataset with [] measurements, collected from individuals grouped into untreated and treated groups. The data is stored in the attached csv file. the dataset has a larger number of columns (i.e. measurements) than the number of rows (individuals). Can you give me a list of statistical approaches to test whether the treatment has significant impacts on the measurements? Can you also talk about the pros and cons of the list of different statistical approaches? |
| | C2. Suggest machine learning approaches with pros and cons | Can you suggest a list of potential candidate machine learning approaches to analyze this type of data? Please provide a description of the pros and cons of the proposed methods. |
| | C3. Generate code with hyperparameters for selected ML methods | Would you provide Python code that reads the data and implements the various methods you suggested? For each method, would you suggest a potential set of hyperparameters to use based on the dataset provided and others' experience with the model? |
| | C4. Generate and test computational frameworks and provide toy models | I have [data] and would like to test the feasibility of using [an algorithm or approach] to analyze it. Can you propose a computational workflow and provide a toy example of the code? |
| **D. Drafting and Writing** | D1. Drafting a manuscript with instruction on each section | 'I am writing a research article based on what we discussed, and on the attached figures and tables. I would like to structure the paper as Introduction, Methods, Results and Discussion. In the Introduction, I would like to focus on []. In the Methods, I would like to include a description of the dataset, the experimental procedure, list all the methods I used to analyze data. In the results, I would like to have several sections: [], [] and [], and cite the figures I generated [with description of the figures]. In the Discussion, I would like to discuss []. Can you provide a draft of the paper with approximately [6000] words?' |
| | D2. Revising a manuscript | 'Can you proofread my manuscript (to be published as a research article in the field of []), correct grammar mistakes and revise it to be suitable for a [] audience? Please highlight the revisions you made.' |

\* [] denotes places where users shall add their own words reflecting what they would like to achieve.

**Table 2. Enhancing Computational Research with LLMs*.**

| Tasks | How LLMs Can Help | Example Prompts |
|---|---|---|
| **1. Automating Code Generation** | - Generate code snippets from natural language descriptions | *"Generate Python code to load a CSV file, remove rows with null values, and display the first 5 rows."* |
| **2. Facilitating Data Analysis & Visualization** | - Propose scripts for exploratory data analysis<br>- Recommend appropriate visualizations (e.g., histograms, scatter plots)<br>- Summarize data with descriptive stats | *"Create Python code using pandas and matplotlib to compute basic descriptive statistics and plot a histogram of the 'age' column."* |
| **3. Assisting in Advanced Computational Tasks** | - Provide scaffolding for simulations and machine learning pipelines<br>- Suggest hyperparameter tuning strategies | *"Provide TensorFlow code to build and train a neural network on the MNIST dataset, including guidance on hyperparameter tuning."* |
| **4. Debugging and Code Optimization** | - Identify potential errors or inefficiencies in code<br>- Offer algorithmic improvements | *"Review the following Python script and suggest improvements to optimize it for faster performance on large datasets."* |
| **5. Enhancing Reproducibility & Accessibility** | - Generate well-documented, standardized code<br>- Integrate best practices for version control | *"Generate a Python script for data cleaning with detailed inline comments and guidelines for Git-based version control."* |
| **6. Bridging Interdisciplinary Gaps** | - Translate complex code into more accessible language<br>- Convert scripts between different programming languages<br>- Explain sophisticated algorithms to non-experts | *"Explain the logic behind this R script in simple terms and provide an equivalent Python version for the same statistical analysis."* |

**\* Important Caveats and Considerations:** While LLMs offer considerable benefits, the code generated by LLMs should be reviewed and tested by researchers to ensure it is accurate, efficient, and tailored to the specific research context. Over-reliance on automated tools increases the risk of errors. We note that this is also true, for example, of the blind use of statistical packages.